\documentclass[runningheads]{llncs}

\usepackage{eccv}

\usepackage{eccvabbrv}

\usepackage{graphicx}
\usepackage{booktabs}

\usepackage[accsupp]{axessibility}  

\usepackage{hyperref}

\usepackage{orcidlink}
\usepackage{amsmath} 
\usepackage{amssymb}
\usepackage{kotex}
\usepackage{multirow}
\usepackage[normalem]{ulem}
\usepackage{wrapfig}

\usepackage[table]{xcolor}

\useunder{\uline}{\ul}{}

\newcommand{\figQualSD}{
\begin{figure*}[!t]
\centering
\includegraphics[width=0.99\textwidth]{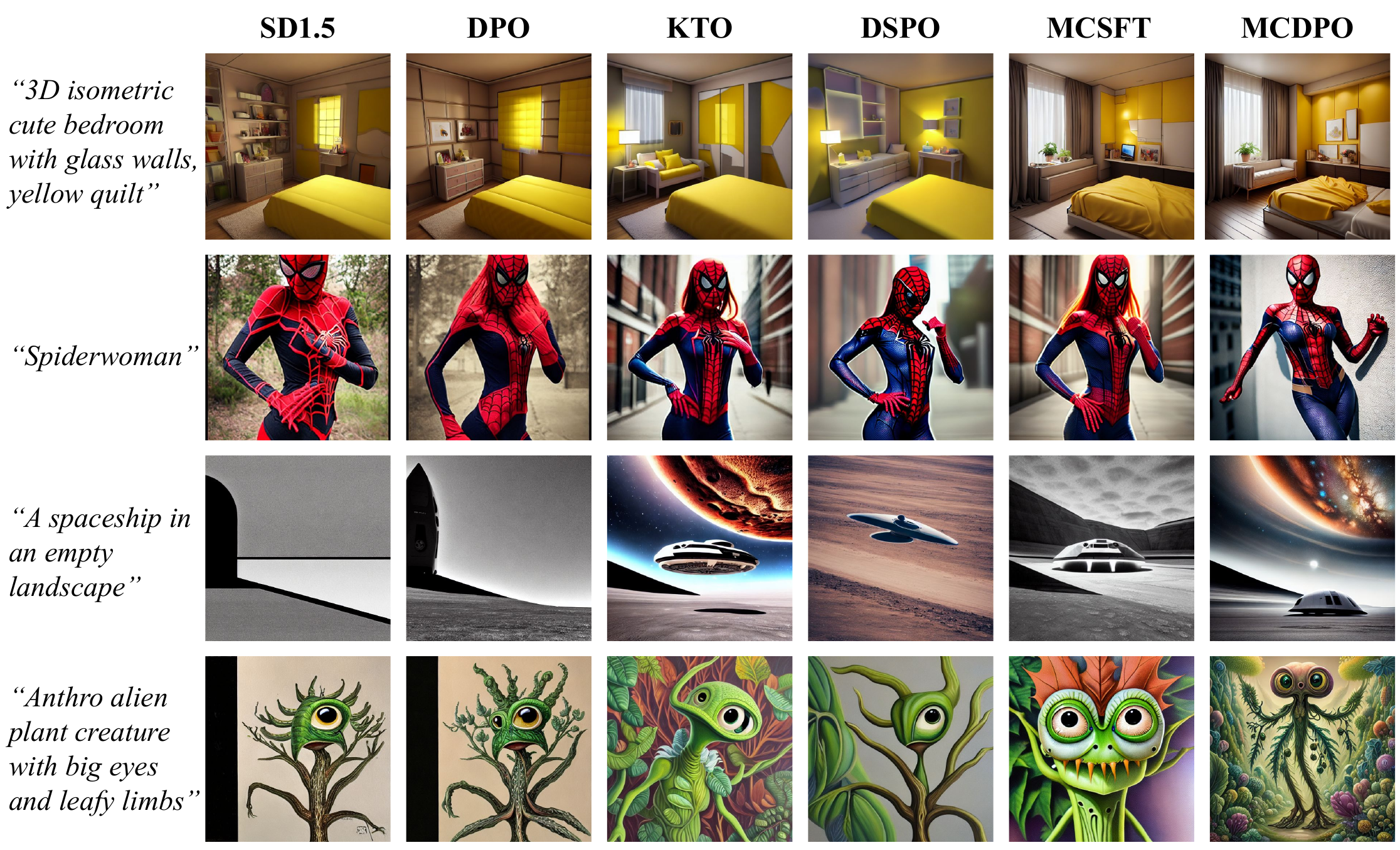}
\vspace{-2mm}
\caption{Qualitative results of SD1.5.}
\vspace{-4mm}
\label{fig:qual_sd15}
\end{figure*}
}

\newcommand{\figQualSDXL}{
\begin{figure*}[!t]
\centering
\includegraphics[width=0.99\textwidth]{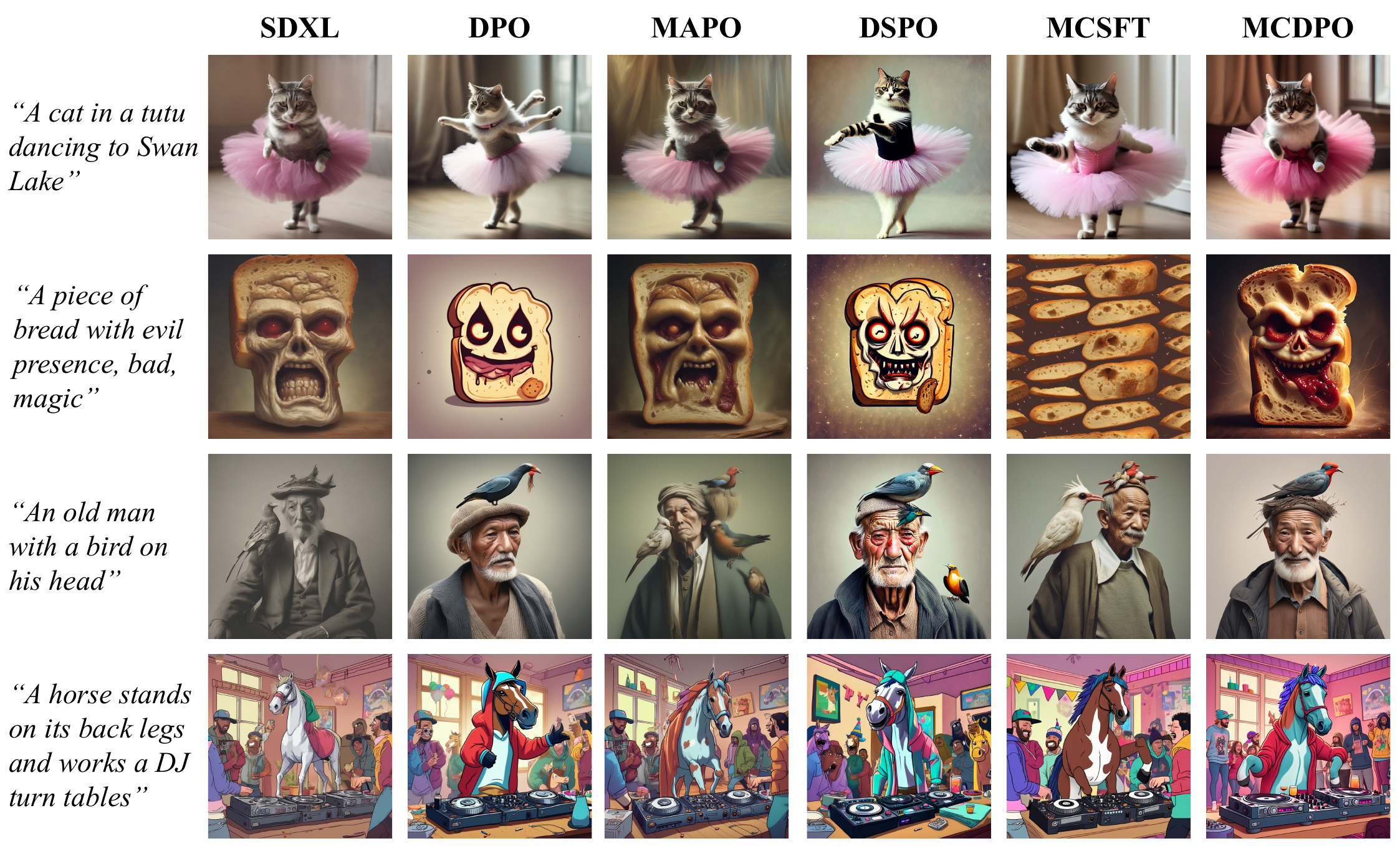}
\vspace{-2mm}
\caption{Qualitative results of SDXL.}
\vspace{-6mm}
\label{fig:qual_sdxl}
\end{figure*}
}

\newcommand{\tabmainSD}{
\begin{table*}[!b]
\centering
\caption{Main Results on SD1.5 (win rates vs. SD1.5 baseline). * indicates the model checkpoint released by the authors. \textbf{Bold} and \underline{underline} indicate the best and second-best results, respectively. We will use the same notation throughout the paper.}
\vspace{-2mm}
\resizebox{0.82\textwidth}{!}{
\begin{tabular}{l|l|cccccc|c}
\hline
Datasets                      & Models & PickScore      & Aesthetic      & HPSv2          & CLIP           & ImageReward  & MPS  & Average        \\ \hline
\multirow{5}{*}{PickV2}       & Diff. DPO*    & 75.52          & 65.08          & 70.28          & {\ul 57.80}           & 64.44          & 67.33    & 66.74         \\
                              & Diff. KTO*    & 73.40          & 71.24          & 82.24          & 56.64          & 76.04          & 69.30  & 71.48         \\
                              & DSPO   & 76.36          & 75.88          & 81.92          & \textbf{59.16}          & 77.36          & 68.80  & 73.24         \\
                              & \cellcolor{gray!25}MCSFT  & \cellcolor{gray!25}{\ul 78.24}          & \cellcolor{gray!25}{\ul 78.16}          & \cellcolor{gray!25}{\ul 89.60}          & \cellcolor{gray!25}57.36          & \cellcolor{gray!25}{\ul 79.20}      & \cellcolor{gray!25}{\ul 69.53}    & \cellcolor{gray!25}{\ul 75.35}          \\
                              & \cellcolor{gray!25}MCDPO  & \cellcolor{gray!25}\textbf{86.20} & \cellcolor{gray!25}\textbf{91.88} & \cellcolor{gray!25}\textbf{93.44} & \cellcolor{gray!25}57.64          & \cellcolor{gray!25}\textbf{82.92} & \cellcolor{gray!25}\textbf{76.75} & \cellcolor{gray!25}\textbf{81.47} \\ \hline
\multirow{5}{*}{PartiPrompts} & Diff. DPO*    & 66.72          & 60.72          & 64.58          & 53.49          & 62.62      &64.69    & 62.14          \\
                              & Diff. KTO*    & 65.74          & 69.85          & {\ul 80.14}          & {\ul 55.82}    & 72.48          & 63.86 & 67.98          \\
                              & DSPO   & {\ul 68.07}          & {\ul 75.18}          & 79.47          & \textbf{56.67} & {\ul 72.85} & {\ul 65.56}          & {\ul 69.63}          \\
                              & \cellcolor{gray!25}MCSFT  & \cellcolor{gray!25}60.42          & \cellcolor{gray!25}68.75          & \cellcolor{gray!25}69.18          & \cellcolor{gray!25}48.49          & \cellcolor{gray!25}65.63 & \cellcolor{gray!25}60.88          & \cellcolor{gray!25}62.22          \\
                              & \cellcolor{gray!25}MCDPO  & \cellcolor{gray!25}\textbf{80.94} & \cellcolor{gray!25}\textbf{92.52} & \cellcolor{gray!25}\textbf{86.46} & \cellcolor{gray!25}50.65          & \cellcolor{gray!25}\textbf{76.35} & \cellcolor{gray!25}\textbf{72.52} & \cellcolor{gray!25}\textbf{76.57} \\ \hline
\multirow{5}{*}{HPDv2}        & Diff. DPO*    & 76.80          & 66.60          & 70.90          & \textbf{56.70}          & 62.50          & 66.66 & 66.69          \\
                              & Diff. KTO*    & 75.55          & 74.40          & 86.35          & 53.20          & 78.20        & 69.17  & 72.81          \\
                              & DSPO   & {\ul 78.10}          & 78.40          & {\ul 86.55}          & {\ul 55.45}          & {\ul 79.80}        & {\ul 70.25}   & {\ul 74.75}          \\
                              & \cellcolor{gray!25}MCSFT  & \cellcolor{gray!25}73.25          & \cellcolor{gray!25}{\ul 78.75}          & \cellcolor{gray!25}81.40          & \cellcolor{gray!25}50.60          & \cellcolor{gray!25}74.45         & \cellcolor{gray!25}70.17 & \cellcolor{gray!25}71.43          \\
                              & \cellcolor{gray!25}MCDPO  & \cellcolor{gray!25}\textbf{87.75} & \cellcolor{gray!25}\textbf{94.25} & \cellcolor{gray!25}\textbf{93.00} & \cellcolor{gray!25}51.75          & \cellcolor{gray!25}\textbf{83.50} & \cellcolor{gray!25}\textbf{76.75} & \cellcolor{gray!25}\textbf{81.16} \\ \hline
\end{tabular}}
\label{tab:mainsd}
\vspace{-2mm}
\end{table*}
}

\newcommand{\tabmainSDXL}{
\begin{table*}[!h]
\centering
\caption{Main Results on SDXL (win rates vs. SDXL baseline).}
\vspace{-2mm}
\resizebox{0.82\textwidth}{!}{
\begin{tabular}{l|l|cccccc|c}
\hline
Datasets                       & Models    & PickScore      & Aesthetic            & HPSv2          & CLIP           & ImageReward    & MPS      & Average         \\ \hline
\multirow{5}{*}{PickV2}       & Diff. DPO* & {\ul 71.60}     & 49.20                & 72.92    & \textbf{61.24} & 68.64  & {\ul 60.36}               & {\ul 63.99}           \\
                              & MAPO*      & 50.65          & {\ul 56.00} & {\ul 80.00}          & 59.15          & 74.19       & 52.26         & 62.02           \\
                              & DSPO      & 68.80     & 44.76                & 71.80          & {\ul 60.60}     & {\ul 74.24}   & 44.77       & 60.82     \\
                              & \cellcolor{gray!25}MCSFT     & \cellcolor{gray!25}50.72          & \cellcolor{gray!25}37.24                & \cellcolor{gray!25}72.36          & \cellcolor{gray!25}59.12          & \cellcolor{gray!25}66.20         & \cellcolor{gray!25}45.96        & \cellcolor{gray!25}55.26           \\
                              & \cellcolor{gray!25}MCDPO     & \cellcolor{gray!25}\textbf{74.92} & \cellcolor{gray!25}\textbf{63.40}           & \cellcolor{gray!25}\textbf{85.84} & \cellcolor{gray!25}58.16    &\cellcolor{gray!25}\textbf{77.72}   & \cellcolor{gray!25}\textbf{64.31}    & \cellcolor{gray!25}\textbf{70.72}  \\ \hline
\multirow{5}{*}{PartiPrompts} & Diff. DPO*       & 64.09          & 54.66                & 67.34          & \textbf{58.21} & 68.44          & {\ul 62.70}      & 62.57          \\
                              & MAPO*      & 50.20          & \textbf{68.07}  & {\ul 76.47}          & 52.38          & 69.73     & 53.10           & 61.65          \\
                              & DSPO      & {\ul 64.58}    & 56.92                & 71.26          & {\ul 56.67}    & \textbf{78.49} & 53.49 & {\ul 63.56}    \\
                              & \cellcolor{gray!25}MCSFT     & \cellcolor{gray!25}46.69          & \cellcolor{gray!25}46.32                & \cellcolor{gray!25}71.51    & \cellcolor{gray!25}55.27          & \cellcolor{gray!25}64.28        & \cellcolor{gray!25}51.96        & \cellcolor{gray!25}56.81          \\
                              & \cellcolor{gray!25}MCDPO     & \cellcolor{gray!25}\textbf{72.06} & \cellcolor{gray!25}{\ul 66.67}          & \cellcolor{gray!25}\textbf{82.41} & \cellcolor{gray!25}55.56    & \cellcolor{gray!25}{\ul 73.47}      & \cellcolor{gray!25}\textbf{64.01}    & \cellcolor{gray!25}\textbf{69.03} \\ \hline
\multirow{5}{*}{HPDv2}        & Diff. DPO*       & 67.80           & 54.85                & 72.10           & \textbf{55.20}  & 67.85    & {\ul 57.53}            & 62.55           \\
                              & MAPO*      & 50.80          & {\ul 56.55}          & {\ul 84.35}           & 54.25          & 70.75            & 48.35     & 60.84           \\
                              & DSPO      & {\ul 70.30}     & 50.35                & 78.80           & 54.75          & {\ul 73.85} & 53.75         & {\ul 63.63}     \\
                              & \cellcolor{gray!25}MCSFT     & \cellcolor{gray!25}51.85          & \cellcolor{gray!25}40.10                 & \cellcolor{gray!25}78.30           & \cellcolor{gray!25}54.75          & \cellcolor{gray!25}66.95     & \cellcolor{gray!25}46.98           & \cellcolor{gray!25}56.48           \\
                              & \cellcolor{gray!25}MCDPO     & \cellcolor{gray!25}\textbf{76.40}  & \cellcolor{gray!25}\textbf{69.25}       & \cellcolor{gray!25}\textbf{90.60}  & \cellcolor{gray!25}{\ul 54.85}    & \cellcolor{gray!25}\textbf{75.20}    & \cellcolor{gray!25}\textbf{69.26}    & \cellcolor{gray!25}\textbf{72.59} \\ \hline
\end{tabular}}
\label{tab:mainsdxl}
\vspace{-4mm}
\end{table*}
}

\begin{document}

\title{Multi-dimensional Preference Alignment by Conditioning Reward Itself} 


\author{Jiho Jang\inst{1}\orcidlink{0009-0004-4824-436X} \and
Jinyoung Kim\orcidlink{0000-0002-9106-2922} \and
Kyungjune Baek\inst{2}\orcidlink{0009-0004-0755-1072}$^\star$ \and
Nojun Kwak\inst{1}\orcidlink{0000-0002-1792-0327}\thanks{Co-corresponding Author}}

\authorrunning{J. Jang et al.}

\institute{Seoul National University \and
Sejong University\\
\email{\{geographic,nojunk\}@snu.ac.kr}, \email{kyungjune.baek@sejong.ac.kr}}

\maketitle

\begin{abstract}
Reinforcement Learning from Human Feedback has emerged as a standard for aligning diffusion models. 
However, we identify a fundamental limitation of the standard DPO formulation when optimizing multi-dimensional preferences: by relying on the Bradley-Terry model, it aggregates heterogeneous evaluation axes (e.g., aesthetic quality and semantic alignment) into a single scalar reward.
This aggregation creates a reward conflict where the model is forced to unlearn desirable features of a specific dimension if they appear in a globally non-preferred sample. To address this issue, we propose Multi Reward Conditional DPO (MCDPO) which resolves reward conflicts by introducing a disentangled Bradley-Terry objective. MCDPO explicitly injects a preference outcome vector as a condition during training, which allows the model to learn the correct optimization direction for each reward axis independently within a single network.
Extensive experiments on Stable Diffusion 1.5 and SDXL demonstrate that MCDPO achieves superior performance on benchmarks even with 18\% and 3\% training data respectively.
Notably, our conditional framework enables dynamic and multiple-axis control at inference time using Classifier Free Guidance to amplify specific reward dimensions without additional training or external reward models.
  \keywords{DPO \and Diffusion \and Multi-preference}
\end{abstract}

\section{Introduction}
\label{sec:intro}

Reinforcement Learning from Human Feedback (RLHF)~\cite{ouyang2022rlhf}
has emerged as a key component in large-scale language model (LLM) training~\cite{liu2024deepseek,yang2025qwen3technicalreport,openai2025gptoss120bgptoss20bmodel}, 
and has recently been extended to generative image models~\cite{xu2023imagereward,lee2023aligning,black2023training}. Among these, Diffusion-DPO~\cite{wallace2024ddpo} applies Direct Preference Optimization (DPO)~\cite{rafailov2023dpo} to diffusion models using only binary win-lose labels via the Bradley-Terry (BT) model.

However, while a single image sample can possess various evaluation axes such as aesthetic quality, prompt fidelity, and safety, the BT model relies solely on a single win-lose label for the entire sample. This formulation makes it difficult to control or reflect specific, multi-dimensional preference axes during training.
This fundamental conflict arises from the standard BT model's formulation, which models global preference as a linear combination of multiple reward axes. 
\begin{wrapfigure}{r}{0.5\textwidth}
\vspace{-2mm}
\centering
\includegraphics[width=0.99\linewidth]{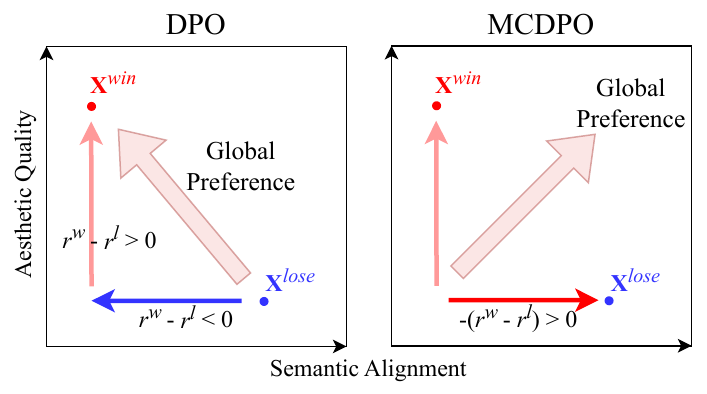}
\vspace{-7mm}
\caption{DPO suffers from reward conflicts ($r^w-r^l < 0$) when global preference contradicts a specific dimension. MCDPO resolves this by disentangling axes to learn the correct direction for each dimension independently ($-(r^w-r^l) > 0$).}
\vspace{-7mm}
\label{fig:concept}
\end{wrapfigure}
Consequently, when presented with a sample pair that is a global \textit{win} but a local \textit{lose} on a specific dimension, 
the model is forced to learn in the opposite direction of the intended preference for that dimension.
Prior works address this through parameter merging~\cite{wang2024conditional}, gradient regularization~\cite{lou2024spo}, SFT-based in-context reward modeling~\cite{yang2024rewards}, or filtering to conflict-free pairs~\cite{lee2025calibrated,gu2025paretohqd}, but none resolve the fundamental conflict within the DPO loss.

To resolve the issue, we propose Multi-reward Conditional Direct Preference Optimization (MCDPO), which injects a per-dimension preference outcome vector as a condition during the DPO loss calculation, transforming problematic reward-conflict pairs into a powerful supervision signal for disentanglement within a single network (See \Cref{fig:concept}). The key idea is a disentangled Bradley-Terry objective that models preferences independently for each dimension by explicitly encoding the win, lose, or tie status for each axis, ensuring every dimension is optimized in its correct direction.

By resolving the reward conflict, we experimentally demonstrate that MCDPO outperforms baselines and forms robust, efficient feature representations for each alignment axis.
Notably, MCDPO trained solely on conflicting pairs surpasses standard DPO trained on randomly sampled data, where standard DPO actively degrades. This confirms that our method transforms reward conflicts from obstacles into a powerful learning signal.
Furthermore, our conditional framework enables dynamic, multi-axis control at inference time by leveraging CFG~\cite{ho2022cfg} to amplify the score function towards higher-reward outcomes without additional training or external reward models.

In summary, the main contributions are as follows:
\vspace{-2mm}
\begin{itemize}
    \item We propose the disentangled BT objective and its practical implementation, Multi-reward Conditional DPO (MCDPO), to address the known reward conflict problem in BT-based DPO. Our method uses a preference outcome vector as a condition to disentangle reward axes.
    
    \item We experimentally show that MCDPO achieves superior performance and sample efficiency, learning robust representations for each alignment axis.
    
    \item We showcase MCDPO's conditional structure, which enables dynamic, multi-axis control during inference using Classifier-Free Guidance.
\end{itemize}

\section{Background}
\label{sec:background}

\subsection{Diffusion Models}

Diffusion models are a class of generative models that learn to generate data by reversing a gradual noising process.~\cite{ho2020denoising} The forward process progressively adds Gaussian noise to data $\mathbf{x}_0$ over $T$ timesteps:
\begin{equation}
q(\mathbf{x}_t|\mathbf{x}_{t-1}) = \mathcal{N}(\mathbf{x}_t; \sqrt{1-\beta_t}\mathbf{x}_{t-1}, \beta_t\mathbf{I}),
\end{equation}
where $\beta_t$ is a variance schedule. The reverse process learns to denoise by training a network $\epsilon_\theta$ to predict the noise at each timestep. The training objective is:
\begin{equation}
\mathcal{L} = \mathbb{E}_{t, \mathbf{x}_0, \epsilon}\left[\|\epsilon - \epsilon_\theta(\mathbf{x}_t, t, c)\|^2\right],
\end{equation}
where $c$ represents conditioning information (e.g., text prompt), and $\epsilon \sim \mathcal{N}(0, \mathbf{I})$ is the added noise to $\mathbf{x}_0$.

At inference, sampling starts from pure noise $\mathbf{x}_T \sim \mathcal{N}(0, \mathbf{I})$ and iteratively denoises to generate $\mathbf{x}_0$. The learned score function $\nabla_{\mathbf{x}} \log p_\theta(\mathbf{x}_t|c)$ guides this reverse process~\cite{song2020score}. 
Classifier-Free Guidance (CFG)~\cite{ho2022cfg} strengthens conditioning by modifying the score function:
\begin{small}
\begin{equation}
\nabla_{\mathbf{x}} \log p_\theta(\mathbf{x}_t) + \lambda_{\text{cfg}} \left(\nabla_{\mathbf{x}} \log p_\theta(\mathbf{x}_t|c) - \nabla_{\mathbf{x}} \log p_\theta(\mathbf{x}_t)\right),
\end{equation}
\end{small}
where $\lambda_{\text{cfg}}$ controls guidance strength. By amplifying the score difference, CFG enhances adherence to the conditioning signal $c$.

\subsection{DPO in Diffusion Models}



Direct Preference Optimization (DPO)~\cite{rafailov2023dpo} offers a simplified approach to align generative models with human preferences by directly optimizing policy models without requiring explicit reward models. DPO leverages the Bradley-Terry (BT) model~\cite{bradley1952rank} to model preference probabilities from pairwise comparisons. Given a pair of samples $\mathbf{x}^w$ (preferred) and $\mathbf{x}^l$ (less preferred) conditioned on prompt $c$, the BT model defines:
\begin{equation}
p_{\text{BT}}(\mathbf{x}^w > \mathbf{x}^l | c) = \sigma(r(\mathbf{x}^w, c) - r(\mathbf{x}^l, c)),
\end{equation}
where $\sigma(\cdot)$ is the sigmoid function and $r(\mathbf{x}, c)$ is a reward function.

A key insight of DPO is that the reward can be expressed analytically in terms of the learned policy $p_\theta$ and a reference policy $p_{\text{ref}}$:
\begin{equation}
r(\mathbf{x}, c) = \beta \log \left(\frac{p_\theta(\mathbf{x}|c)}{p_{\text{ref}}(\mathbf{x}|c)}\right) + \beta \log Z(c),
\end{equation}
where $\beta$ is a KL regularization parameter and $Z(c)$ is the partition function. This formulation allows DPO to optimize preferences directly through the policy without training a separate reward model.

For diffusion models~\cite{wallace2024ddpo}, the DPO objective operates at the denoising transition level. Given a preference (win-lose) pair $(\mathbf{x}^w, \mathbf{x}^l)$, the loss function is:
\begin{footnotesize}
\begin{equation}
\begin{split}
\mathcal{L}_{\text{DPO}} =& -\mathbb{E} \log \sigma \left( \beta \log \frac{p_\theta(\mathbf{x}^w_t|\mathbf{x}^w_{t+1}, c)}{p_{\text{ref}}(\mathbf{x}^w_t|\mathbf{x}^w_{t+1}, c)} - \beta \log \frac{p_\theta(\mathbf{x}^l_t|\mathbf{x}^l_{t+1}, c)}{p_{\text{ref}}(\mathbf{x}^l_t|\mathbf{x}^l_{t+1}, c)} \right) \\
=& -\mathbb{E} \log \sigma (-\beta T \omega_t\ (||\epsilon^w - \epsilon_\theta(\mathbf{x}_t^w,t)||^2 - ||\epsilon^w - \epsilon_\text{ref}(\mathbf{x}_t^w,t)||^2 \\
&\ \ \ \ \ \ \ \ \ \ \ \ \ \ \ \ \ \ \ \  - (||\epsilon^l - \epsilon_\theta(\mathbf{x}_t^l,t)||^2 - ||\epsilon^l - \epsilon_\text{ref}(\mathbf{x}_t^l,t)||^2)
),
\end{split}
\end{equation}
\end{footnotesize}
where the expectation is over timesteps $t$ and noisy samples from the diffusion forward process. This formulation trains the diffusion model to increase the likelihood of preferred samples while decreasing that of less preferred ones, relative to the reference policy.






In practice, generated samples are evaluated along multiple distinct axes, such as aesthetic quality, semantic alignment, and safety. However, the standard BT model aggregates these into a single scalar reward by a linear combination of different dimensions:
\begin{equation}
r(\mathbf{x}, c) = \sum_{i=1}^D w_i r_i(\mathbf{x}, c),
\end{equation}
where $r_i$ represents the reward for dimension $i$ and $w_i$ its weight. This leads to a fundamental reward conflict: when $x^w$ is preferred globally but performs worse than $x^l$ on a specific dimension $j$, the DPO loss may inadvertently degrade that dimension as shown in \Cref{fig:concept} (Left). 

Existing studies attempt to address or bypass this issue. In the LLM field, prominent approaches include: (1) dynamically combining expert model parameters based on user-specified reward weights~\cite{wang2024conditional}, (2) 
training sequentially with regularization that encourages gradient alignment with previously learned preferences~\cite{lou2024spo}, and (3) avoiding DPO conflicts by using supervised fine-tuning (SFT) that incorporates reward scores directly in the prompt~\cite{yang2024rewards}.
In the diffusion domain, a common strategy to avoid this conflict is to generate images until the samples that are dominant across all axes as preferred~\cite{lee2025calibrated,zhang2024learning}.
Nevertheless, these approaches fail to resolve the fundamental conflict within the DPO loss formulation itself. By either training only on non-conflicting pairs (which is sample-inefficient) or using alternative objectives (like SFT or regularization), they rely on a limited learning signal rather than directly solving the ambiguity in DPO.
In this work, we aim to analyze the problem and resolve the limitations.

\section{Method}
\label{sec:method}
We first analyze the reward conflict and propose a disentangled BT objective (\Cref{subsec:reward_conflict}), then introduce MCDPO as its practical single-network implementation (\Cref{subsec:mcdpo}), address gradient domination via reward dropout (\Cref{subsec:reward_drop}), and show how our conditional framework enables test-time reward control (\Cref{subsec:testtime}).

\subsection{Multi-reward Disentanglement}
\label{subsec:reward_conflict}
As discussed in \Cref{sec:background}, the standard BT model collapses multi-dimensional
preferences into a single scalar preference:
\begin{equation}
    p_{BT}(\mathbf{x}^w>\mathbf{x}^l|c)
    =
    \sigma\!\left[\sum_i w_i\big(r_i(\mathbf{x}^w,c)-r_i(\mathbf{x}^l,c)\big)\right].
\end{equation}
This formulation leads to a reward conflict when a sample $\mathbf{x}^w$ is globally preferred over $\mathbf{x}^l$, but is worse than $x^l$ on a particular dimension $j$, (i.e.,
$r_j(\mathbf{x}^w,c) < r_j(\mathbf{x}^l,c)$).
Because the standard DPO loss optimizes only for the global win, it inadvertently compels the model to learn in the opposite direction for dimension $j$, actively degrading that specific quality.
To resolve this, we first introduce a disentangled Bradley-Terry objective. The core idea is to model the preference for each dimension. We explicitly introduce a pairwise preference outcome vector $\gamma(\mathbf{x},\mathbf{y}) \in \{-1,0,1\}^D$, where each entry indicates whether $\mathbf{x}$ wins, loses, or ties against $\mathbf{y}$ on reward axis $i$, \ie, $\gamma_i(x,y) = \text{sign}(r_i(x) - r_i(y))$.

Using $\gamma$, we define the disentangled Bradley-Terry $p^\perp_{BT}$ target:
\footnotesize
\begin{equation}
\begin{split}
    p^\perp_{BT}(\mathbf{x}^w>\mathbf{x}^l \mid c,\gamma(\mathbf{x}^w,\mathbf{x}^l))
    =
    \sigma\!\left[
    \sum_i w_i \gamma_i(\mathbf{x}^w,\mathbf{x}^l)
    \big(r_i(\mathbf{x}^w,c)-r_i(\mathbf{x}^l,c)\big)
    \right].
\end{split}
\label{eq:disentangle_bt}
\end{equation}
\normalsize

Eq.~\eqref{eq:disentangle_bt} specifies the desired preference probability once the per-dimension outcomes are known.
This formulation ensures each dimension is optimized in the correct direction. In particular, if the $j$-th dimensional preference is inverted (e.g., $\mathbf{x}^l$ wins on aesthetics), $\gamma_j$ becomes negative, which flips the sign for that dimension's contribution. By doing so, we can resolve the sign ambiguity in conflicting pairs and provide correct supervision across all dimensions.

\subsection{Multi-reward Conditional DPO}
\label{subsec:mcdpo}
Because standard DPO is structurally limited to modeling a single reward axis, modeling multiple axes would require training multiple separate implicit reward diffusion models.
\begin{equation}
    r_i(\mathbf{x},c) = \beta \log\left(\frac{p_i(\mathbf{x}|c)}{p_\text{ref}(\mathbf{x}|c)}\right) + \beta \log Z(c)
\end{equation}
this approach would require optimizing a loss:
\footnotesize
\begin{equation}
    L_M = -\mathbb{E}\ \log\sigma\left[\beta \sum_iw_i\gamma_i \left( \log\frac{p_i(\mathbf{x}^w|c)}{p_{\text{ref}}(\mathbf{x}^w|c)} - \log\frac{p_i(\mathbf{x}^l|c)}{p_{\text{ref}}(\mathbf{x}^l|c)}\right)\right],
    \label{eq:lm}
\end{equation}
\normalsize 
which has two major drawbacks. First, it does not fundamentally resolve conflicts, as each model is still trained on pairs where its dimension may conflict with others.
Second, it requires training $D$ separate models which is computationally prohibitive.
The first issue can be addressed by incorporating the per-dimension preference indicator $i$ from our disentangled BT objective \eqref{eq:disentangle_bt}, and the separate models can be further consolidated into a single network conditioned on $i$:
\footnotesize
\begin{equation}
    L_M' = -\mathbb{E}\ \log\sigma\left[\beta \sum_iw_i\gamma_i \left( \log\frac{p(\mathbf{x}^w|c,i)}{p_{\text{ref}}(\mathbf{x}^w|c)} - \log\frac{p(\mathbf{x}^l|c,i)}{p_{\text{ref}}(\mathbf{x}^l|c)}\right)\right].
    \label{eq:lm_prime}
\end{equation}
\normalsize 
This formulation effectively resolves reward conflicts, as conditioning the model on $i$ ensures that each specific quality dimension is optimized in its intended direction.
However, it remains computationally prohibitive. The summation inside the sigmoid requires evaluating the log-likelihood ratio independently for each dimension $i$, necessitating $D$ separate forward-backward passes per training step.



To resolve this, we propose Multi-reward Conditional DPO (MCDPO), an efficient solution that models $p_{BT}^\perp$~\eqref{eq:disentangle_bt} using a single conditional diffusion model.
Following DPO, we interpret the policy/reference log-ratio as an implicit reward, but make it conditional on the pairwise outcome vector $\gamma$:
\begin{equation}
    r_\theta(\mathbf{x},c,\gamma)
    =
    \beta \log \frac{p_\theta(\mathbf{x}\mid c,\gamma)}{p_{\mathrm{ref}}(\mathbf{x}\mid c)} + \beta \log Z(c,\gamma).
\end{equation}
This induces a conditional Bradley-Terry probability
\begin{equation}
\begin{split}
    \hat p_\theta^\perp&(\mathbf{x}^w>\mathbf{x}^l \mid c,\gamma)
    =
    \sigma\!\left(
    r_\theta(\mathbf{x}^w,c,\gamma)-r_\theta(\mathbf{x}^l,c,\gamma)
    \right)
    \\&=
    \sigma\!\left(
    \beta \log \frac{p_\theta(\mathbf{x}^w\mid c,\gamma)}{p_{\mathrm{ref}}(\mathbf{x}^w\mid c)}
    -
    \beta \log \frac{p_\theta(\mathbf{x}^l\mid c,\gamma)}{p_{\mathrm{ref}}(\mathbf{x}^l\mid c)}
    \right).        
    \end{split}
    \label{eq:conditional_bt}
\end{equation}


Under sufficient model capacity and an expressive conditioning mechanism,
the conditional policy $p_\theta(\mathbf{x}|c,\gamma)$ can, in principle, learn
a distinct implicit reward function for each value of $\gamma \in \{-1,0,1\}^D$.

Let $r_{\text{eff}}(\mathbf{x}, c, \gamma) \triangleq \sum_i w_i \gamma_i\, r_i(\mathbf{x}, c)$ be the effective reward under condition $\gamma$. 
By construction, the effective reward difference
satisfies:
\footnotesize
\begin{equation}
r_{\text{eff}}(\mathbf{x}^w, c, \gamma) - r_{\text{eff}}(\mathbf{x}^l, c, \gamma)
= \sum_i w_i |r_i(\mathbf{x}^w, c) - r_i(\mathbf{x}^l, c)| \geq 0,
\label{eq:consistency}
\end{equation}
\normalsize
which guarantees that the training label ($\mathbf{x}^w$ preferred) is consistent with the effective reward $r_{\text{eff}}(\mathbf{x}, c, \gamma)$ 
under the Bradley-Terry model for every $\gamma$. Since each fixed $\gamma$ reduces $L_{MC}$ (Eq.~\eqref{eq:lmc}) to a standard DPO objective with ground-truth reward $r_{\text{eff}}$, the DPO optimality result~\cite{rafailov2023dpo} directly implies that at the optimum $\theta^*$, the  induced reward satisfies:
\begin{equation}
r_{\theta^*}(\mathbf{x}, c, \gamma) = \sum_i w_i \gamma_i\, r_i(\mathbf{x}, c) + g(c, \gamma),
\label{eq:reward_decomp}
\end{equation}
where $g(c, \gamma)$ is independent of $\mathbf{x}$ and thus cancels in the pairwise difference
$r_{\theta^*}(\mathbf{x}^w, c, \gamma) - r_{\theta^*}(\mathbf{x}^l, c, \gamma)$.
Substituting this into Eq.~\eqref{eq:conditional_bt} recovers the disentangled target
$p^\perp_{BT}$ in Eq.~\eqref{eq:disentangle_bt}, establishing
$\hat{p}^\perp_{\theta^*} = p^\perp_{BT}$. 

Intuitively, the training dynamics naturally drive the model toward this decomposition. Consider a conflicting pair where $\gamma=[\text{aesthetic: $+1$},\ \text{semantic: $-1$}]$. The loss requires the model to increase the likelihood of $\mathbf{x}^w$ along the aesthetic axis while \emph{decreasing} it along the semantic. The same pair also appears with the flipped condition $\gamma=[\text{aesthetic: $-1$},\ \text{semantic: $+1$}]$ via our final symmetric loss Eq.~\eqref{eq:lmc}, which imposes the opposite constraint. To satisfy both simultaneously, the model must learn to modulate its implicit reward for each dimension independently as a function of $\gamma$, which is precisely the structure captured by Eq.~\eqref{eq:reward_decomp}:
While Eq.~\eqref{eq:reward_decomp} may not hold exactly in practice due to the finite capacity of the conditioning mechanism,~\Cref{tab:implicit} empirically validates that our architecture approximates this decomposition well. The single MCDPO model achieves implicit reward accuracy comparable to specialist models trained on individual dimensions.
\footnotesize
\begin{equation}
\begin{split}
    L_{MC} = -\mathbb{E}\ \Biggl[ \log\sigma \Biggl( &\beta \log\frac{p(\mathbf{x}^w|c,\gamma^{wl})}{p_{\text{ref}}(\mathbf{x}^w|c)} - \beta \log\frac{p(\mathbf{x}^l|c,\gamma^{wl})}{p_{\text{ref}}(\mathbf{x}^l|c)} \Biggl)\\
    + \log\sigma\Biggl(&\beta \log\frac{p(\mathbf{x}^l|c,\gamma^{lw})}{p_{\text{ref}}(\mathbf{x}^l|c)} - \beta \log\frac{p(\mathbf{x}^w|c,\gamma^{lw})}{p_{\text{ref}}(\mathbf{x}^w|c)} \Biggl)\Biggl],
\end{split}
\label{eq:lmc}
\end{equation}
\normalsize
where $\gamma^{wl}$ and $\gamma^{lw}$ denote $\gamma(x^w,x^l)$ and $\gamma(x^l,x^w)$, respectively. To feed $\gamma$ into the model, we devise a reward conditioning module described in \Cref{subsec:contextawaremodule}.

\subsection{Mitigating Gradient Domination}
\label{subsec:reward_drop}
While our $L_{MC}$ objective~\Cref{eq:lmc} successfully disentangles reward axes, naively training with it can lead to gradient domination.
This occurs when dimensions that are easier to learn suppress the learning signals for harder dimensions, leading to unbalanced optimization.

Under the reward decomposition at optimum (Eq.~\eqref{eq:reward_decomp}), analyzing the gradient of $L_{MC}$ reduces to an equivalent analysis of $L_M'$ (Eq.~\eqref{eq:lm_prime}), since the conditional policy $p_\theta(\mathbf{x}|c,\gamma)$ decomposes into the same per-dimension structure. We can write the gradient of the loss as follows:
\footnotesize
\begin{equation}
\begin{split}
    &\nabla_\theta L_M' = (\sigma (z)-1)\nabla_\theta z ,\\
    \nabla_\theta z = \sum_i\beta w_i &(\nabla_\theta \log p_\theta(\mathbf{x}^w|c,i) - \nabla_\theta \log p_\theta(\mathbf{x}^l|c,i)),
\end{split}
\end{equation}
\normalsize
where $z$ aggregates the per-dimension log-likelihood ratios. We can find two sources of the issue.
First, if any single dimension becomes highly confident, $\sigma(z) \rightarrow 1$, the global gradient goes to zero; therefore, the entire training converges to the local optima. Second, within $\nabla_\theta z$, if the model 
becomes highly discriminative for dimension $i$, which causes a large gradient norm for that dimension; it can dominate the sum and overwhelm gradients from other dimensions.

To address this, we introduce dimensional reward dropout. During training, we randomly drop individual reward dimensions by setting their condition $\gamma_i = 0$. By doing so, we can zero out the contribution of the dropped dimension to both the $\sigma(z)$ term and the summed gradient $\nabla_{\theta}z$. This simple technique prevents any single dimension from consistently dominating the training signal, ensuring balanced optimization across all axes.

\subsection{Test-time Reward Optimization}
\label{subsec:testtime}
A significant advantage of our conditional framework is that it enables dynamic, multi-axis reward optimization at inference time. Our training objective implicitly guides the model to represent two distinct distributions, including the distribution of preferred samples $p(\mathbf{x}|c,\gamma^w)$ and the distribution of non-preferred samples $p(\mathbf{x}|c,\gamma^l)$:
\small
\begin{equation}
    p(\mathbf{x}|c,\gamma^w) = p_{\text{ref}}(\mathbf{x}|c)\exp(r^w(\mathbf{x},c)/\beta)/Z(c).
    \label{eq:implicit}
\end{equation}
\normalsize

This structure is a natural fit for Classifier-Free Guidance (CFG). By computing the difference between the score functions for the ``win'' and ``lose'' conditions, we can derive the gradient of the implicit reward difference ($r^w - r^l$):
\small
\begin{equation}
    \nabla_\mathbf{x} \log\ p(\mathbf{x}|c,\gamma^w)-\nabla_\mathbf{x} \log\ p(\mathbf{x}|c,\gamma^l) = \nabla_\mathbf{x}( r^w- r^l)/\beta.
\end{equation}
\normalsize
The gradient can be incorporated into the standard CFG sampling as follows:
\small
\begin{equation}
\nabla_\mathbf{x} \log\ p(\mathbf{x}|\gamma^l) + \lambda_{cfg}(\nabla_\mathbf{x} \log\ p(\mathbf{x}|c,\gamma^w)-\nabla_\mathbf{x} \log\ p(\mathbf{x}|\gamma^l)).
\end{equation}
\normalsize
By doing so, we can steer the generation process away from the non-preferred ($\gamma^l$) and toward the preferred ($\gamma^w$). This enables fine-grained, dynamic control, allowing a user to specify a weighted combination of preference gradients at test time without requiring any external reward models.

\subsection{Context-aware Reward Conditioning}
\label{subsec:contextawaremodule}
We encode the preference outcome vector $\gamma$ as natural language tokens (e.g., ``win tie'', meaning win for the first (\eg, aesthetic) score and tie for the second (\eg, semantic alignment) via the text encoder to get an initial embedding $e_\gamma$. This embedding is then injected into the model.

First, the U-Net latent $z_t$ is updated using parallel cross-attention (CA) blocks with the text prompt $c_{\text{prompt}}$ and the final context-aware reward (CAR) embedding $c_{\text{reward}}$:
\begin{equation}
    z_t' = \text{CA}(z_t, c_{\text{prompt}}) + \lambda \cdot \text{CA}'(z_t, c_{\text{reward}}),
    \label{eq:ipadapter}
\end{equation}
where $z_t'$ is the updated latent, $\text{CA}'$ is a copy of the pretrained CA module~\cite{ye2023ipadapter}, and $\lambda$ is a weighting scalar, which we set to 1 in all our experiments.

For dimensions requiring both text and image context (e.g., semantic alignment), the embedding $c_{\text{reward}}$ is computed using a context-aware conditioning module. This module refines the initial $\gamma$ embedding $e_\gamma$ by attending to both the text prompt and the image latent sequentially:
\begin{equation}
    h_{\text{SA}} = \text{SA}(e_\gamma, c_{\text{prompt}}), \ h_{\text{CA}} = \text{CA}''(h_{\text{SA}}, z_t), \
    c_{\text{reward}} = \text{MLP}(h_{\text{CA}}),
\label{eq:car}
\end{equation}
where SA, $\text{CA}''$, and MLP are zero-initialized to preserve pretrained knowledge. This final $c_{\text{reward}}$ embedding is then fed into~\Cref{eq:ipadapter} as the input to $\text{CA}'$.

\noindent\textbf{Training Initialization} The modules introduced above ($\text{CA}'$ and the context-aware reward conditioning) are newly added parameters that are zero-initialized to preserve pretrained knowledge. As a result, the conditioning signal $\gamma$ has no effect at initialization.
Just as SFT before DPO in LLMs~\cite{rafailov2023dpo} helps the model
learn basic instruction-following before preference optimization,
we perform a supervised fine-tuning stage (MCSFT) that trains the conditioning
modules to reconstruct preferred samples $\mathbf{x}^w$ given $\gamma^w$
with a frozen U-Net, allowing the model to learn basic condition-following
before the MCDPO alignment begins.
We emphasize that MCSFT serves solely as initialization, as shown in~\cref{tab:mainsdxl,tab:conflict}. MCSFT alone is insufficient and the subsequent MCDPO alignment stage is critical for final performance.


\section{Experiments}
\label{sec:experiments}

\subsection{Experimental Setup}
\noindent\textbf{Training Dataset.}
Following the previous works, we adopt Pick-a-Pic v2~\cite{kirstain2023pick}, containing 1M human-preference image pairs as the training dataset.
We augment each pair with four model-based reward dimensions: aesthetic quality (Aesthetic Predictor~\cite{laionaes}), semantic alignment (CLIP~\cite{radford2021learning}), overall quality (HPSv2~\cite{wu2023human}), and prompt alignment (PickScore~\cite{kirstain2023pick}).
For our method, the 5-dimensional preference outcome vector $\gamma$ is then constructed by encoding the win/lose/tie status for the original human label alongside these four proxy reward dimensions by discretized into 100 bins (same bin = ties) after normalization on training dataset. We use 18\% of the data ($\sim$180K pairs) for StableDiffusion1.5 (SD1.5)~\cite{Rombach_2022_CVPR} and 3\% ($\sim$30K pairs) for SDXL~\cite{podell2023sdxl} while baselines use the full dataset.

\tabmainSD
\tabmainSDXL

\noindent\textbf{Configuration.} Training proceeds in two phases: (1) MCSFT pre-training and (2) MCDPO alignment . The total computational time, including the prior MCSFT pretraining, was 16 and 32 A100 GPU hours for SD1.5 and SDXL, respectively. Other implementation details are provided in the Appendix.

\noindent\textbf{Evaluation.}
To evaluate MCDPO's effectiveness, we compare against existing baselines trained on the Pick-a-Pic v2 dataset: Diffusion-DPO~\cite{wallace2024ddpo}, Diffusion-KTO~\cite{lialigning}, MAPO~\cite{hong2024mapo}, and DSPO~\cite{zhu2025dspo}. Following standard convention, we test text-to-image generation on Pick-a-Pic v2 test set~\cite{kirstain2023pick}, PartiPrompts~\cite{yu2022scaling}, and HPDv2 test set~\cite{wu2023human}, measuring performance with PickScore, LAION Aesthetic Score, HPSv2, CLIP, ImageReward~\cite{xu2023imagereward}, and MPS~\cite{zhang2024learning}.

\subsection{Main Results}
As shown in \Cref{tab:mainsd}, MCDPO demonstrates dominant performance on the SD1.5, significantly outperforming all baselines (Diffusion-DPO, Diffusion-KTO, and DSPO) across every test set. We observe substantial gains in both PickScore and HPSv2. The most notable improvement is in the aesthetic score, which shows 
\begin{wraptable}{r}{0.5\textwidth}
  \centering
  \vspace{-6mm}  
  \caption{\textbf{Fair comparison across training regimes.} All models are trained on an identical 18\% data budget. MCDPO significantly outperforms all baselines, including filtered and conflict-only regimes, demonstrating its robust conflict resolution.}
    \label{tab:conflict}
    \resizebox{1.0\linewidth}{!}{
\begin{tabular}{l|cccccc|c} \hline
               & Pick & Aes  & HPS  & CLIP & IR   & MPS  & Avg  \\ \hline
(1) DPO            & 75.1 & 62.6 & 71.7 & {\ul 60.4} & 62.6 & 68.1 & 66.7 \\
(2) DPO merged     & 73.4 & 63.5 & 70.0 & 56.1 & 62.0 & 67.9 & 63.8 \\
(3) CDPO merged    & 63.4 & {\ul 84.8} & 71.7 & 32.3 & 61.1 & 65.0 & 63.1 \\
(4) Filtered DPO   & 83.6 & 84.1 & 85.7 & \textbf{62.8} & {\ul 79.2} & {\ul 76.6} & {\ul 78.7} \\
(5) MCSFT (full)   & 78.1 & 78.3 & 81.7 & 55.4 & 77.0 & 68.7 & 73.2 \\
(6) MCDPO          & \textbf{86.2} & \textbf{91.8} & \textbf{93.4} & 57.6 & \textbf{82.9} & \textbf{76.7} & \textbf{81.1} \\
(7) Conflict DPO   & 62.0 & 57.0 & 60.4 & 56.4 & 56.6 & 61.9 & 59.0 \\
(8) Conflict MCDPO & {\ul 85.2} & 83.2 & {\ul 89.4} & 56.0 & 74.0 & 74.3 & 77.0 \\ \hline
\end{tabular}}
  \vspace{-14pt}
\end{wraptable}
a massive performance gain over existing methods. Furthermore, MCDPO achieves consistent gains on ImageReward and MPS, metrics not used as a direct reward dimension, indicating strong generalization capabilities.
This significant advantage is maintained on SDXL, as detailed in \Cref{tab:mainsdxl}. MCDPO again outperforms all baselines on PickScore, HPSv2, and MPS, achieving the highest average score by a large margin.


While MCDPO achieves substantial improvements across human-preference 
metrics, CLIP scores are slightly lower than standard DPO. This reflects a natural multi-objective trade-off, where the moderate CLIP enables dramatic gains in aesthetics (91.9\%), HPSv2 (93.4\%), ImageReward (82.9\%), and MPS (76.7\%). Importantly, \Cref{tab:singledim_inf} confirms that MCDPO has learned a disentangled CLIP representation that can be amplified at test-time. In~\Cref{tab:singledim_inf}, targeting CLIP alone yields 59.0\%, and Aes-CLIP temporal scheduling, which assigns aesthetic `tie' for the first 25 steps and `win' for the remaining 25, further recovers the gap to 59.6\% while preserving the highest overall average (81.5\%).

To ensure a fair comparison across various training regimes, we evaluate our method using an identical 18\% subset of the Pick-a-Pic v2 dataset. \Cref{tab:conflict} compares eight distinct configurations to demonstrate how effectively MCDPO resolves reward conflicts. (1) Standard DPO on randomly sampled pairs, (2)
merged checkpoints of DPO models trained independently on each reward axis — same supervision, (3) merged checkpoints of Conditional DPO (CDPO) models independently trained on each axis — same supervision and parameters (4) DPO-Filtered on conflict-free pairs only, (5) full MCSFT training (unfreezing the U-Net after the initial MCSFT phase) — same supervision and parameters (6) our proposed MCDPO on randomly sampled pairs, (7) DPO trained strictly on conflict-only pairs, and (8) MCDPO trained strictly on conflict-only pairs.

\figQualSD
\figQualSDXL

Notably, MCDPO surpasses DPO-Filtered, which explicitly avoids conflicting signals by discarding 82\% of training data, demonstrating that MCDPO successfully transforms conflict pairs from obstacles into highly effective training signals. Furthermore, we observe that full MCSFT training alone actually degrades performance, confirming that the MCDPO alignment objective is indispensable for stable optimization.
To explicitly demonstrate that our model fundamentally resolves reward conflicts, we compare the models trained exclusively on conflicting pairs ((7) and (8)).
When restricted to conflict-only data, standard DPO suffers a substantial relative performance degradation of 11.5\% compared to its randomly sampled baseline. In contrast, MCDPO experiences a minor drop of only 5.1\%. In fact, Conflict MCDPO (Avg 77.0) trained solely on conflicting pairs surpasses standard DPO on randomly sampled data (Avg 66.7) by a large margin, confirming that MCDPO does not merely tolerate reward conflicts but actively leverages them as a stronger supervision signal.

Remarkably, the conflict-trained MCDPO maintains high performance even when conditioned on ``all-win'' across every dimension, a scenario never encountered during training. We provide further analysis in~\Cref{fig:reward_traj} and agreement ratio over reward dimensions in appendix.

Furthermore, we visualize the qualitative result of MCDPO against baseline methods on both SD1.5 and SDXL, in~\Cref{fig:qual_sd15} and~\Cref{fig:qual_sdxl}, respectively.
Across both models, our method demonstrates a clear superiority in generating images that are more aesthetically pleasing and detailed.

\subsection{Analysis}


\noindent\textbf{Multi-dimensional Learning.} \Cref{tab:singledim_dpo,tab:singledim} compare MCDPO against specialist models trained on a single reward axis,
using both the standard DPO objective (\Cref{tab:singledim_dpo}) and our Conditional DPO paradigm with identical architecture and hyperparameters (\Cref{tab:singledim}).

Remarkably, the MCDPO model not only achieves the highest average score but also outperforms every individual specialist model on its own specialized metric, with the sole exception of the CLIP score in~\Cref{tab:singledim_dpo}. For example, the MCDPO model (Aesthetic win-rate 91.8\%) significantly surpasses the model trained exclusively on `Aesthetic' rewards (Aesthetic win-rate 88.1\%) in~\Cref{tab:singledim}.

We conjecture that MCDPO mitigates potential training reward overfitting by leveraging cross-dimensional references to learn robust features rather than train-specific shortcuts. For example, one dimension may exploit spurious cues that improve training reward but harm semantics. The CLIP dimension can provide a semantic reference that helps mitigate this.



\begin{table}[t]
\centering
\begin{minipage}[t]{0.49\linewidth}
    \centering
    \caption{\textbf{Standard DPO specialists.}}
    \label{tab:single_dpo}
    \vspace{-2mm}
    \resizebox{0.99\linewidth}{!}{
\begin{tabular}{l|cccccc|c}
\hline
          & Pick     & Aes     & HPS         & CLIP          & IR   & MPS           & Avg       \\ \hline
Human     & {\ul 75.1}    & 62.6          & 71.7          & \textbf{60.4} & {\ul 62.6}    & {\ul 68.1}    & {\ul 66.7}    \\
Pick & 72.6          & 63.4          & {\ul 72.0}    & {\ul 59.2}    & 57.6          & 66.6          & 65.2          \\
Aes & 69.8          & {\ul 66.3}    & 66.4          & 56.4          & 59.5          & 66.0          & 64.1          \\
HPS     & 69.8          & 63.9          & 70.1          & 54.5          & 59.5          & 65.9          & 64.0          \\
CLIP      & 69.3          & 62.4          & 67.7          & 58.5          & 58.9          & 63.2          & 63.3          \\
Merged    & 73.4          & 63.5          & 70.0          & 56.1          & 62.0          & 67.9          & 63.8          \\ \hline
MCDPO     & \textbf{86.2} & \textbf{91.8} & \textbf{93.4} & 57.6          & \textbf{82.9} & \textbf{76.7} & \textbf{81.1} \\ \hline
\end{tabular}
\label{tab:singledim_dpo}
    }
\end{minipage}
\hfill 
\begin{minipage}[t]{0.49\linewidth}
    \centering
    \caption{\textbf{Cond. DPO specialists.}}
    \label{tab:singledim}
    \vspace{-3mm}
    \resizebox{0.99\linewidth}{!}{
        \begin{tabular}{l|cccccc|c} \hline
         & Pick & Aes  & HPS  & CLIP & IR & MPS   & Avg  \\ \hline
        Human  & 79.0 & 79.8 & 86.5 & 46.8 & 75.9 & 73.2 & 73.5 \\
        Pick   & {\ul 80.7} & 82.8 & {\ul 88.5} & 49.0 & 76.8 & {\ul 74.7} & {\ul 75.4} \\
        Aes    & 79.6 & {\ul 88.1} & 87.2 & 49.1 & 76.2 & 72.5 & {\ul 75.4} \\
        HPS    & 77.8 & 81.8 & 88.0 & {\ul 50.2} & {\ul 77.6} & 73.3 & 74.7 \\
        CLIP   & 75.3 & 79.3 & 84.5 & 49.3 & 76.0 & 70.9 & 72.5 \\
        Merged   & 63.4 & 84.8 & 71.7 & 32.3 & 61.1 & 65.0 & 63.1 \\ \hline
        MCDPO    & \textbf{86.2} & \textbf{91.8} & \textbf{93.4} & \textbf{57.6} & \textbf{82.9} & \textbf{76.7} & \textbf{81.1} \\ \hline
        \end{tabular}
    }
\end{minipage}
\vspace{-5mm}
\end{table}

\begin{figure}[t]
    \centering
    
    \begin{minipage}[c]{0.48\textwidth}
        \centering
        \captionof{table}{Test-time reward optimization by guiding the MCDPO model towards specific target dimensions.}
        \label{tab:singledim_inf}
        \resizebox{\linewidth}{!}{
            \begin{tabular}{l|cccccc|c} \hline
                & Pick    & Aes           & HPS           & CLIP          & IR       & MPS     & Avg           \\ \hline
                Human & {\ul 78.6} & 79.0          & 83.3          & 55.2          & 73.7      & 67.7    & 72.9          \\
                Pick  & 78.1       & 79.0          & 85.1          & {\ul 57.6}    & {\ul 78.4} & 69.9   & 74.7          \\
                Aes   & 77.3       & \textbf{95.8} & 79.1          & 49.3          & 73.1     & 68.8     & 73.8          \\
                HPS   & 77.0       & 79.5          & {\ul 85.6}    & \textbf{59.0} & 77.8     & {\ul 71.1}     & {\ul 74.9}    \\
                CLIP  & 76.4       & 76.5          & 85.1          & \textbf{59.0} & 75.0    & 67.8      & 73.3          \\ \hline
                Aes-CLIP Cont.   & 88.0 & 90.6 & 91.2 & 59.6 & 81.0 & 78.4 & 81.5 \\
                ALL   & \textbf{86.2} & {\ul 91.8}    & \textbf{93.4} & {\ul 57.6}    & \textbf{82.9} & \textbf{76.7} & \textbf{81.1} \\ \hline
            \end{tabular}
        }
    \end{minipage}
    \hfill 
    \begin{minipage}[c]{0.48\textwidth}
        \vspace{0pt}
        \centering
        \includegraphics[width=\linewidth]{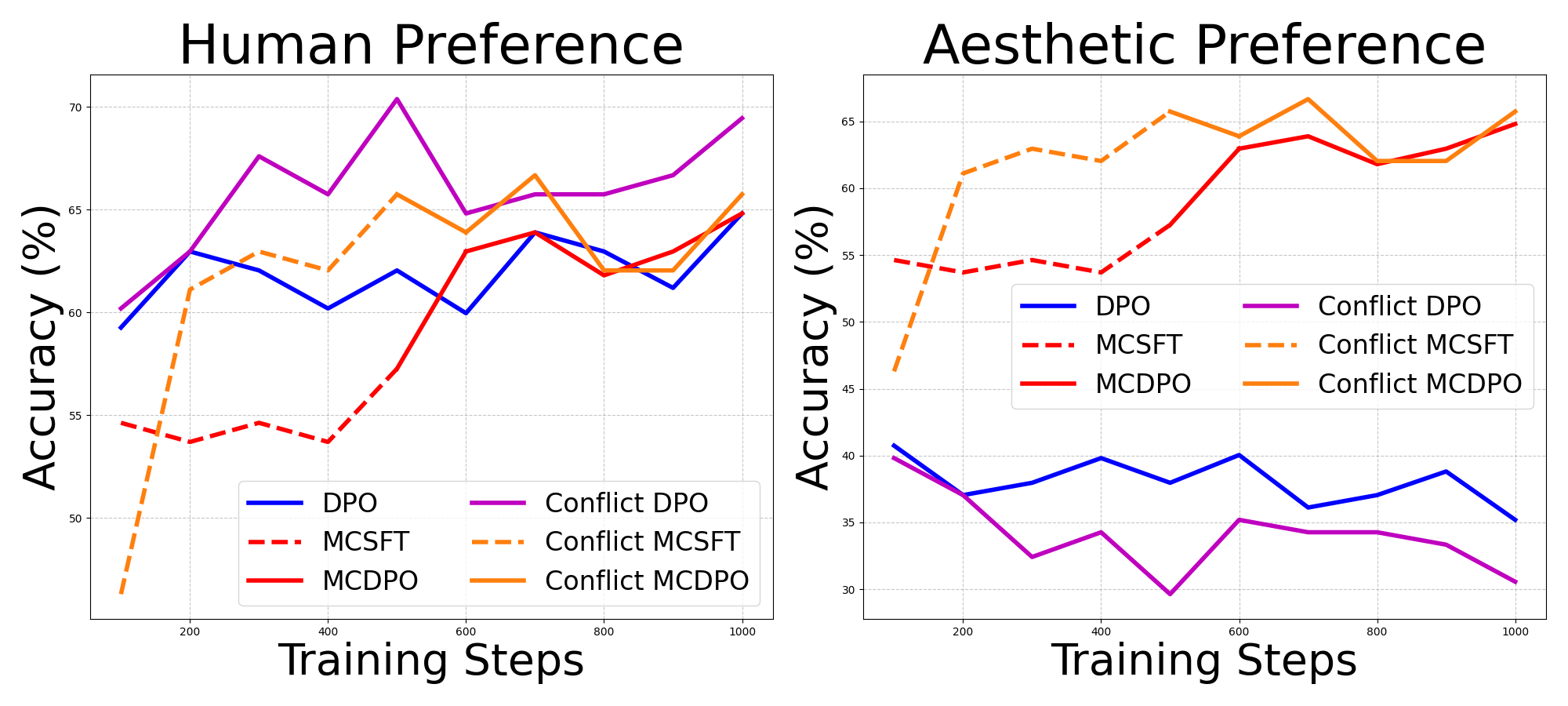}
        \captionof{figure}{Implicit reward accuracy on Human-Aesthetic conflict-only pairs.}
        \label{fig:reward_traj}
    \end{minipage}
    \vspace{-5mm}
\end{figure}

\noindent\textbf{Multi-dimensional Inference.}
As demonstrated in~\Cref{tab:singledim_inf}, our single model, trained to disentangle reward axes, can perform dynamic preference-enhancing sampling at test-time as described in~\Cref{subsec:testtime}. When steering the sampling to amplify only a specific or multiple dimensions, we observe distinct and controlled behaviors. Critically, when enhancing CLIP, the model achieves 59.0\%, recovering from the all-dimension default (57.6\%) and demonstrating that MCDPO has learned a disentangled CLIP representation that can be amplified when desired.

This demonstrates that MCDPO has successfully learned a disentangled representation of CLIP rewards and can boost this dimension when desired.
Similarly, enhancing Aes (Aesthetic) generates samples with exceptionally high aesthetic scores (95.8\%), validating dimension-specific control.

Notably, when enhancing PickScore and HPSv2, their corresponding scores increase as expected. However, we also observe a concurrent rise in CLIP, ImageReward, and MPS scores.
We attribute this to the nature of PickScore and HPSv2 as holistic metrics that evaluate overall image quality, which naturally correlates with other reward axes.

While coarse axis-level optimization is detailed in the supplementary material, we highlight MCDPO's capability for fine-grained temporal control.
To mitigate the natural trade-off between aesthetics and semantics, we leverage the step-wise frequency properties of diffusion models. Semantic structures primarily form during early low-frequency steps, whereas aesthetic details emerge in later high-frequency steps. By assigning an aesthetic tie for the first 25 steps and a win condition for the remaining 25 steps, our Aes-CLIP Control (\Cref{tab:singledim_inf}) significantly boosts the CLIP score while maintaining the average win rate. We further provide an incremental analysis in Appendix, where adding each reward axis as ``win'' progressively improves the corresponding metric, confirming that each dimension contributes independently to the final performance.


\noindent\textbf{Implicit Reward Modeling}
~\Cref{tab:implicit} demonstrates that MCDPO accurately models implicit reward functions for each dimension. MCDPO achieves comparable accuracy to specialist models trained on individual dimensions, while being significantly more computationally efficient by learning all dimensions within a single network.
Furthermore, our conditional DPO framework simultaneously models both $r^w$ (preferred) and $r^l$ (non-preferred) implicit reward models as described in~\Cref{eq:implicit}. Leveraging both models together yields performance gains over using either model alone. For example, in the Aesthetic dimension, while the win-only model achieves 50.8\% and the lose-only model achieves 53.3\%, combining both models substantially improves accuracy to 66.2\%. Similarly, for Human and PickScore dimensions, despite the lose models showing relatively lower accuracies (42.2\% and 41.0\% respectively), combining them with the win models consistently yields performance improvements.

\noindent\textbf{Reward Conflicts}
To demonstrate that MCDPO actively resolves reward conflicts beyond mere dimensional disentanglement, we plot the implicit rewards during training, on a filtered subset where Human and Aesthetic preferences conflict \Cref{fig:reward_traj}.
While standard DPO and Conflict DPO suffer active aesthetic feature degradation dropping to 35\% and 30\% respectively, MCDPO successfully rectifies the gradient direction via its axis-wise guidance $\gamma$. This isolates the aesthetic dimension from the global penalty and enables convergence towards the ideal disentangled distribution $p_{BT}^\perp$.

\begin{table}[!t]
\centering
\begin{minipage}[t]{0.49\linewidth}
    \centering
    \caption{Implicit reward accuracy comparison between MCDPO and specialist.}
    \label{tab:implicit}
    \vspace{-2mm}
    \resizebox{0.99\linewidth}{!}{
    \begin{tabular}{l|ccccc} \hline
               & Human & Pick & Aes  & HPS  & CLIP \\ \hline
    Specialists & 57.8  & 61.0 & 67.1 & 58.2 & 59.9 \\
    MCDPO      & 58.7  & 61.7 & 66.2 & 55.0 & 59.4 \\ \hline
    MCDPO win  & 58.5  & 58.7 & 50.8 & 52.9 & 59.4 \\
    MCDPO lose & 42.2  & 41.0 & 53.3 & 46.8 & 56.8 \\ \hline
    \end{tabular}
    }
\end{minipage}
\hfill 
\begin{minipage}[t]{0.49\linewidth}
    \centering
    \caption{Ablation study of MCDPO components on PickV2.}
    \label{tab:ablation}
    \vspace{-2mm}
    \resizebox{0.95\linewidth}{!}{
        \begin{tabular}{l|cccccc|c} \hline
                        & Pick & Aes & HPS  & CLIP & IR  & MPS & Avg \\ \hline
        MCDPO           & {\ul 86.2}  & 91.8 & \textbf{93.4} & \textbf{57.6} & \textbf{83.0} & {\ul 76.7} & \textbf{81.1} \\ \hline
        reward dropout 0.1   & 85.8  & {\ul 93.5} & {\ul 92.4} & {\ul 55.6} & {\ul 82.4} & 74.8 & {\ul 80.7} \\
        reward dropout 0.0   & \textbf{87.3}  & \textbf{96.0} & 92.1 & 49.2 & 79.2 & \textbf{77.1} & 80.1 \\
        w/o MCSFT         & 53.8  & 79.2 & 55.1 & 48.8 & 55.7 & 70.7 & 60.5 \\
        w/o CAR module & 80.1  & 90.0 & 80.7 & 45.7 & 69.8 & 54.4 & 70.1 \\ \hline
        \end{tabular}
    }
\end{minipage}
\vspace{-4mm}
\end{table}


\noindent\textbf{Component Ablation Study.} We conduct an ablation study to validate the contribution of MCDPO's core components: dimensional reward dropout, the context-aware reward module, and the MCSFT pre-training stage in~\Cref{tab:ablation}.

First, removing the dimensional reward dropout leads to a seemingly 
dominant aesthetic score (95.8\%) but causes a catastrophic collapse in the CLIP score (49.3\%). We believe this is a clear case of the gradient domination problem discussed in \Cref{subsec:reward_drop}. The easier aesthetic dimension appears to saturate the learning signal, preventing the harder CLIP dimension from being optimized. Our reward dropout technique effectively mitigates this issue, ensuring balanced optimization and contributing significantly to the superior overall performance. Applying a reward dropout of just 0.1 significantly mitigates this domination.

Removing MCSFT pre-training causes a drastic performance decline, showing non-initialized conditioning is disruptive. However, MCSFT alone is insufficient. On SDXL, MCSFT performs poorly (Avg 57.1), while the full MCDPO model achieves 72.0 (\Cref{tab:mainsdxl}), proving the MCDPO alignment stage itself is critical.

Finally, ablating the Context-Aware Reward conditioning module Eq.~\eqref{eq:car} causes a severe performance drop in all metrics except for aesthetics. These affected metrics (PickScore, HPS, CLIP, IR) measure alignment between the image and the text prompt. This result confirms our hypothesis: to condition on rewards like semantic alignment effectively, the model must be aware of both the image and the text context, validating the design of our module.




\section{Conclusion}
\label{sec:conclusion}



We introduced Multi Reward Conditional DPO (MCDPO), a framework that resolves the reward conflict in the Bradley-Terry formulation by conditioning on preference outcome vectors, transforming conflicting pairs into robust training signals for disentangled alignment. MCDPO achieves state-of-the-art performance with superior sample efficiency, and enables dynamic multi-dimensional control at inference without additional training or external reward models. One limitation is the reliance on proxy reward models to construct the preference outcome vector, which upper-bounds alignment quality by their accuracy, and introduces preprocessing overhead compared to standard DPO's binary labels.

\section{Acknowledgement}
This work was supported by the Korean Government through the grants from IITP (RS-2021-II211343, RS-2022-II220320, RS-2025-25442338) and NRF (RS-2026-25470591).

\bibliographystyle{splncs04}
\bibliography{main}

\clearpage
\onecolumn
\setcounter{page}{1}
\begin{center}
  {\Large \textbf{Multi-dimensional Preference Alignment by Conditioning Reward Itself} \\ \vspace{0.5em}Supplementary Material \\
  \vspace{1.0em}}
\end{center}

\appendix

\section{Analysis of Reward Correlation and Conflict}
\label{sec:reward_corr}

To provide a quantitative foundation for our paper's core premise, we analyzed the relationships between the reward dimensions used in our training data. The central motivation for MCDPO is the existence of "reward conflict," where different preference axes provide contradictory optimization signals. We computed the Pearson correlation matrix for the five primary reward dimensions across our training set (derived from Pick-a-Pic v2~\cite{kirstain2023pick}) in \Cref{tab:corr}.

\begin{table}[h]
\centering
\vspace{-3mm}
\caption{Pearson correlation matrix of the five reward dimensions used in the training dataset.}
\begin{tabular}{l|ccccc}
\hline
          & Human  & PickScore & Aesthetic & HPSv2  & CLIP    \\ \hline
Human     & 1.0000 & 0.2649    & 0.0097    & 0.0960 & 0.1277  \\
PickScore & 0.2649 & 1.0000    & -0.0501   & 0.4683 & 0.4174  \\
Aesthetic & 0.0097 & -0.0501   & 1.0000    & 0.1083 & -0.0662 \\
HPSv2     & 0.0960 & 0.4683    & 0.1083    & 1.0000 & 0.4325  \\
CLIP      & 0.1277 & 0.4174    & -0.0662   & 0.4325 & 1.0000  \\ \hline
\end{tabular}
\vspace{-3mm}
\label{tab:corr}
\end{table}

This correlation matrix provides direct, quantitative validation for our paper's thesis. The most critical finding is the Aesthetic dimension, which shows a near-zero or slightly negative correlation with all other preference metrics, including Pickscore (-0.0501), CLIP Score (-0.0662), and Human Preference (0.0097). This data proves that optimizing for what a human prefers provides no signal for optimizing aesthetics. This is the ``reward conflict" we identify. A standard DPO model, which aggregates these signals into a single scalar, is forced to average these contradictory gradients, leading to suboptimal performance on some axes. While PickScore, CLIP Score, and HPS Score show moderate positive correlations (0.41-0.46), they are far from 1.0, indicating they capture related but distinct aspects of quality and semantic alignment.

Furthermore, the Human Preference label, which serves as the ground truth alignment target, shows only a weak correlation with any single proxy model (e.g., 0.1277 with CLIP Score and 0.0960 with HPS Score). This finding suggests the complexity of the alignment task, as it indicates that no single proxy serves as a strong substitute for the true, complex human preference. This suggests the model must learn from a combination of noisy, low-correlation, and conflicting signals. These findings collectively suggest that reward conflict is a measurable property of the data, not just a theoretical edge case. This motivates a solution beyond simple aggregation and provides a strong justification for our proposed method, MCDPO, which is explicitly designed to disentangle and optimize these axes independently.

\begin{figure*}
    \centering
    \vspace{-3mm}
    \includegraphics[width=1.0\linewidth]{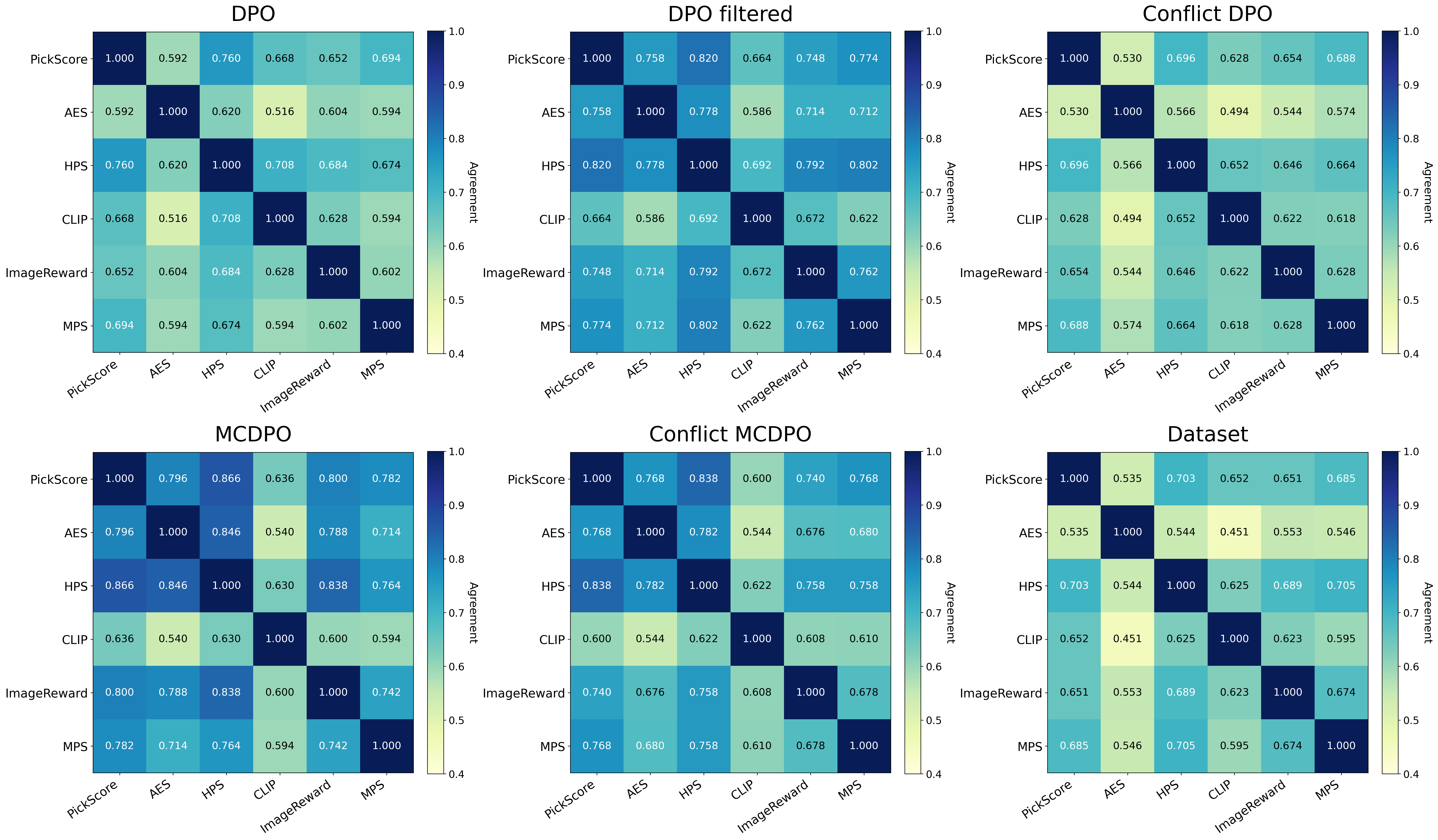}
    \vspace{-2mm}
    \caption{Pairwise preference agreement between reward 
dimensions on generated outputs. Each cell $(i,j)$ represents 
the fraction of generated sample pairs where dimensions $i$ 
and $j$ assign the same preference direction (i.e., both 
prefer $\mathbf{x}^w$ or both prefer $\mathbf{x}^l$).}
    \label{fig:agreement}
    \vspace{-3mm}
\end{figure*}

\Cref{fig:agreement} visualizes the pairwise preference 
agreement on generated outputs. While the correlation matrix 
(\Cref{tab:corr}) characterizes the training data, the 
agreement heatmap reflects how well each trained model resolves 
conflicts at generation time. MCDPO achieves the highest 
overall agreement, 
demonstrating that it aligns multiple axes simultaneously 
rather than trading off between them. Even Conflict MCDPO 
maintains substantially higher agreement than both standard 
DPO and Conflict DPO, confirming that our method transforms 
reward conflicts into coherent optimization signals.

\section{Test-time Reward Axis Interpolation.}
\begin{table}[h]
    \centering
    \caption{Test-time reward axis interpolation on PickV2. Starting from Human-only guidance, each row cumulatively adds one reward axis as ``win''. Each added axis progressively improves its corresponding metric, confirming that MCDPO has learned an independently controllable representation per dimension.}
    \label{tab:test_time_reward_axis}
    \vspace{-2mm}
\begin{tabular}{l|cccccc|c} \hline
                   & Pick & Aes  & HPS  & CLIP & IR   & MPS  & Avg  \\\hline
Human -- (1)       & 78.6 & 79.0 & 83.2 & 55.2 & 73.8 & 74.0 & 74.0 \\
(1) + Pick -- (2)  & 83.1 & 81.5 & 88.4 & 57.2 & 76.9 & 72.0 & 76.5 \\
(2) + Aes -- (3)   & 86.0 & \textbf{94.8} & 88.6 & 49.1 & 79.0 & 77.1 & 79.1 \\
(3) + HPS -- (4)   & \textbf{89.9} & {\ul 94.7} & {\ul 91.3} & 52.0 & 80.3 & {\ul 79.0} & {\ul 81.2} \\
(4) + CLIP (MCDPO) & 86.2 & 91.9 & \textbf{93.4} & {\ul 57.6} & \textbf{82.9} & 76.8 & \textbf{81.5} \\ \hline
Aes-CLIP Control   & {\ul 88.0} & 90.6 & 91.2 & \textbf{59.6} & {\ul 81.0} & \textbf{78.4} & \textbf{81.5} \\ \hline
\end{tabular}
\vspace{-3mm}
\end{table}

We provide an incremental analysis of MCDPO's test-time reward control by progressively adding each reward axis as ``win'' during CFG-based sampling.
As shown in Table~\ref{tab:test_time_reward_axis}, starting from Human-only guidance~(1), adding PickScore~(2) raises Pick from 78.6 to 83.1 and concurrently improves HPS and CLIP.
Introducing the Aesthetic axis~(3) yields a dramatic aesthetic gain (81.5 $\to$ 94.8) but reduces CLIP to 49.1, reflecting the known aesthetic--semantic trade-off.
Adding HPS~(4) partially recovers CLIP (49.1 $\to$ 52.0) while further boosting the overall average.
Finally, adding CLIP restores it to 57.6 and achieves the highest average of 81.5.

This progressive pattern confirms that each reward dimension contributes independently to the final performance, validating that MCDPO learns a disentangled implicit reward per axis rather than a single entangled representation.
Furthermore, the Aes-CLIP temporal control achieves the same overall average (81.5) while attaining the highest CLIP score (59.6), demonstrating that the aesthetic--semantic trade-off can be further mitigated through step-wise scheduling as discussed in Section~3.4.

%

\section{Implementation Details}

Our training process consists of two main phases: Multi-reward Conditional Supervised Fine-Tuning (MCSFT) and Multi-reward Conditional Direct Preference Optimization (MCDPO). We use AdamW~\cite{loshchilov2017decoupled} for SD1.5 and Adafactor~\cite{shazeer2018adafactor} for SDXL following baselines. In the initial MCSFT stage, the parameters of the original U-Net are frozen, and only the conditioning module described in~\Cref{subsec:contextawaremodule} of the main paper is trained. For SD1.5, we train for 500 steps with a batch size of 384. For SDXL, 200 steps with a batch size of 64. Both use a base learning rate of 1e-9 linearly scaled with the batch size and 100 warmup steps. During this stage, we apply a reward condition dropout rate of 0.2 for model-based rewards and 0.0 for Human Preference, along with a text condition dropout rate of 0.2. In the subsequent MCDPO stage, all model parameters, including the U-Net and the conditioning module, are fine-tuned. The conditioning module is initialized from the MCSFT weights. We train for 500 steps with a DPO $\beta$ of 6000 and 100 warmup steps. We apply dimensional reward dropout with rates of 0.15 for CLIP, 0.2 for the other model-based rewards, and 0.0 for Human Preference, corresponding to the default MCDPO configuration in Table~8 of the main paper. For SD1.5, we use a batch size of 384 and a base learning rate of 1e-9. For SDXL, a batch size of 64 and a base learning rate of 5e-9, both linearly scaled with the batch size. The total training time for both stages was 16 A100 80GB GPU hours (2 GPUs $\times$ 6 hours) for SD1.5 and 32 A100 80GB GPU hours (2 GPUs $\times$ 16 hours) for SDXL. For evaluation, we use DDIM sampling with 50 steps. We report each method at its best-performing CFG scale for a fair comparison.

\section{Human Evaluation}
\begin{table}[h]
\centering
\caption{Human Evaluation.}
\begin{tabular}{l|ccccc}
\hline
                    & Base  & DPO   & KTO   & DSPO   & MCDPO  \\ \hline
Mean Rank           & 3.7   & 3.28  & 2.98  & 2.90   & \textbf{2.14}   \\
Top-1 Rate & 5.1\% & 7.1\% & 7.1\% & 13.1\% & \textbf{46.5\%} \\ \hline
\end{tabular}
\end{table}
We conducted a human evaluation on 100 prompts randomly sampled from the PickV2 test set. Human evaluators ranked outputs from all models based on SD1.5.

\section{Additional Ablation Study.}
\begin{table}[h]
    \centering
    \caption{Ablation on CLIP reward dropout rate $p$ on PickV2. 
All other dropout rates are fixed at their default values 
(0.2 for model-based rewards, 0.0 for Human Preference).}
    \label{tab:ablation_clip_dropout}
    \vspace{-2mm}
\begin{tabular}{l|cccccc|c}
\hline
 & Pick & Aes & HPS & CLIP & IR & MPS & Avg \\ \hline
$p=0.0$ & 85.5 & 91.7 & {\ul 93.2} & 57.3 & {\ul 83.3} & 76.2 & 81.2 \\
$p=0.05$ & {\ul 86.4} & 91.1 & 92.9 & \textbf{57.8} & 82.7 & 76.1 & 81.2 \\
$p=0.10$ & 86.2 & \textbf{91.9} & 92.8 & 57.2 & \textbf{83.7} & 76.5 & {\ul 81.4} \\
$p=0.15$ (Ours) & 86.2 & \textbf{91.9} & \textbf{93.4} & {\ul 57.6} & 82.9 & \textbf{76.8} & \textbf{81.5} \\
$p=0.20$ & \textbf{86.6} & {\ul 91.8} & \textbf{93.4} & 57.3 & 83.0 & 76.7 & \textbf{81.5} \\ \hline
\end{tabular}
\end{table}

\noindent\textbf{CLIP dropout} As discussed in Section~3.3, the CLIP dimension is particularly 
susceptible to gradient domination due to its inherent difficulty 
relative to other reward axes such as aesthetics. We therefore 
ablate the CLIP-specific reward dropout rate $p$ while keeping 
all other dropout rates fixed. As shown in 
Table~\ref{tab:ablation_clip_dropout}, performance is relatively 
stable across all values, indicating that our method is not 
sensitive to this hyperparameter. We select $p=0.15$ as it 
achieves the highest average win rate (81.5) while maintaining 
balanced performance across all metrics, particularly on HPS 
(93.4) and MPS (76.8).

\begin{table}[h]
\centering
\caption{Additional Ablation Study. Avg. win-rate on SD1.5.}
\begin{tabular}{l|cccc|ccc|cc}
\hline
\multirow{2}{*}{} & \multicolumn{4}{c|}{Reward CFG Scale}  & \multicolumn{3}{c|}{Num. of Bins} & \multicolumn{2}{c}{Noise Ratio} \\
                  & 1.0  & 3.0  & 5.0  & 7.5  & 10        & 50        & 100      & 10\%           & 20\%           \\ \hline
Avg. Win rate     & 74.5 & 79.1 & 79.7 & 79.8 & 81.2      & 82.0      & 81.9     & 80.6           & 80.7           \\ \hline
\end{tabular}
\end{table}

\noindent\textbf{Discretization Sensitivity} Sensitivity to discretization is reported in Tab. 12, sweeping bin counts. Average win rate remains stable across all settings, confirming robustness. The ternary design follows the BT formulation directly, as $\gamma \in {-1, 0, 1}$ encodes sign($r_i^w - r_i^l$) and preserves the multiplicative structure of Eq. 9.

\noindent\textbf{Partially missing and noisy labels}
Missing labels are handled by setting the outcome entry to `\textit{tie}', masking that dimension from supervision as in reward dropout. For label noise, we randomly flipped each dimension at 10\% and 20\%. Tab. 12 shows robustness to partially noisy labels.

\noindent\textbf{Test-time optimization via CFG}
For the CFG effect, Eq. 21 entangles prompt and reward condition under a single CFG coefficient, obscuring the reward guidance effect. We therefore introduce a disentangled formulation below and sweep reward CFG strength in Tab. 12, where performance consistently improves.

\vspace{-4mm}
\begin{footnotesize}
\begin{equation}
\begin{split}
\nabla_\mathbf{x} \log\ &p(\mathbf{x}|\gamma^{tie}) + \lambda_{cfg}(\nabla_\mathbf{x} \log\ p(\mathbf{x}|c,\gamma^{tie})-\nabla_\mathbf{x} \log\ p(\mathbf{x}|\gamma^{tie})) \\
&+ \lambda_{cfg}^{reward}(\nabla_\mathbf{x} \log\ p(\mathbf{x}|c,\gamma^{w})-\nabla_\mathbf{x} \log\ p(\mathbf{x}|c,\gamma^{l}))
\end{split}
\raisetag{11pt}
\end{equation}
\end{footnotesize}

\section{Additional Qualitative Results}

\begin{figure}[h]
    \centering
    \includegraphics[width=0.9\linewidth]{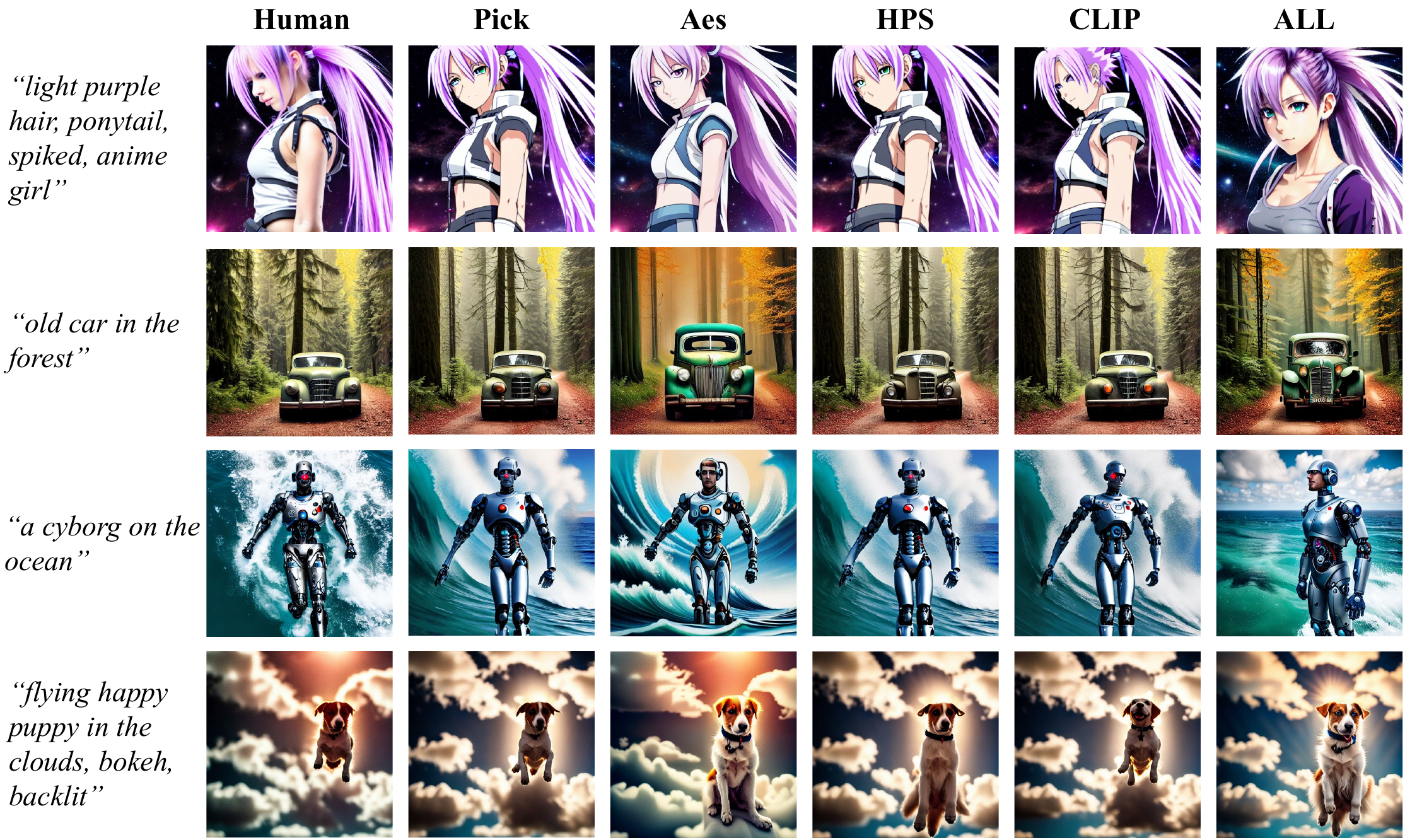}
    \caption{Visualization of single-dimensional inference (\Cref{tab:singledim_inf}).}
    \label{fig:singledim}
\end{figure}

\begin{figure*}
    \centering
    \includegraphics[width=\linewidth]{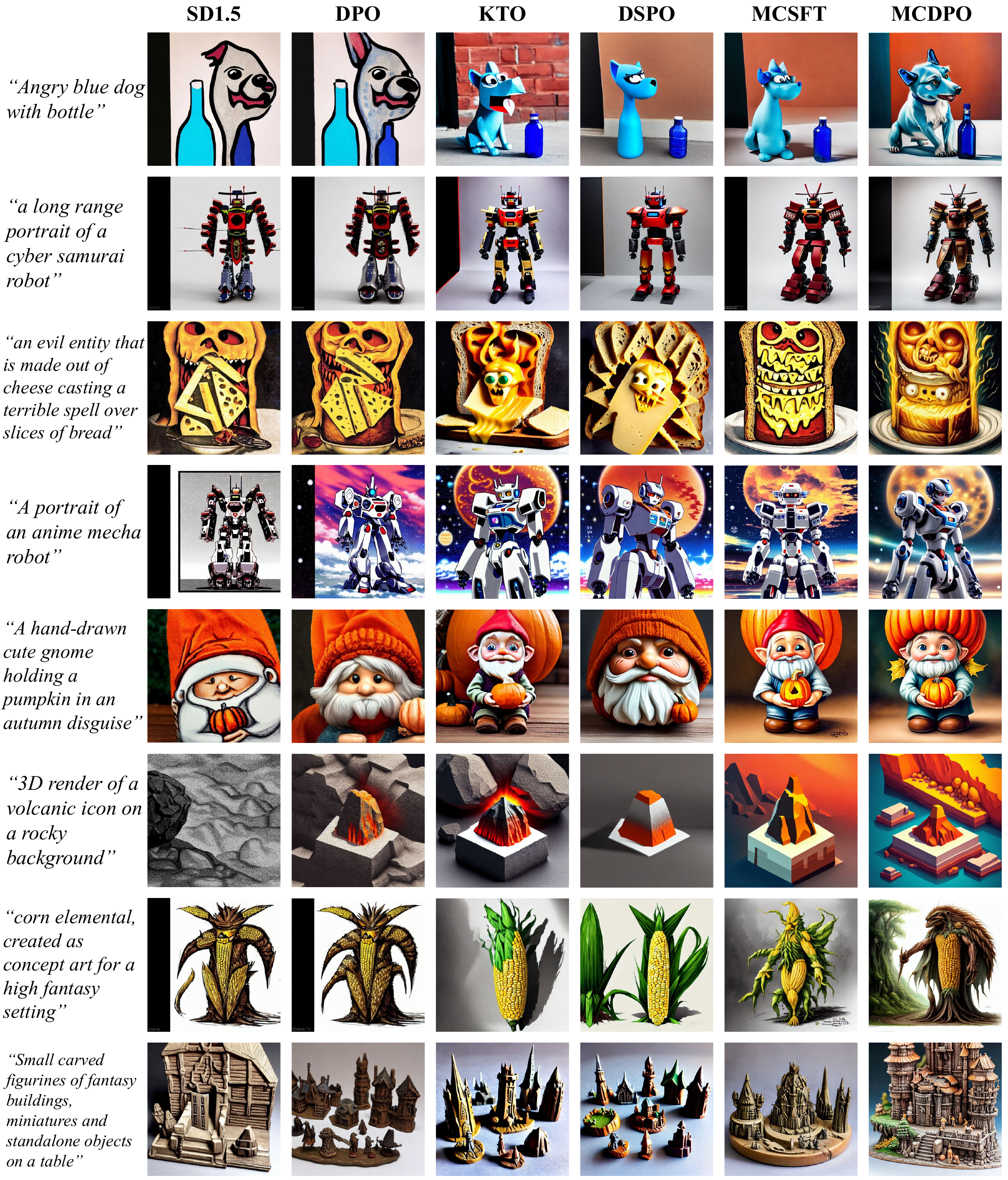}
    \caption{Additional qualitative results of SD15.}
    \label{fig:sd15_supp}
\end{figure*}

\begin{figure*}
    \centering
    \includegraphics[width=\linewidth]{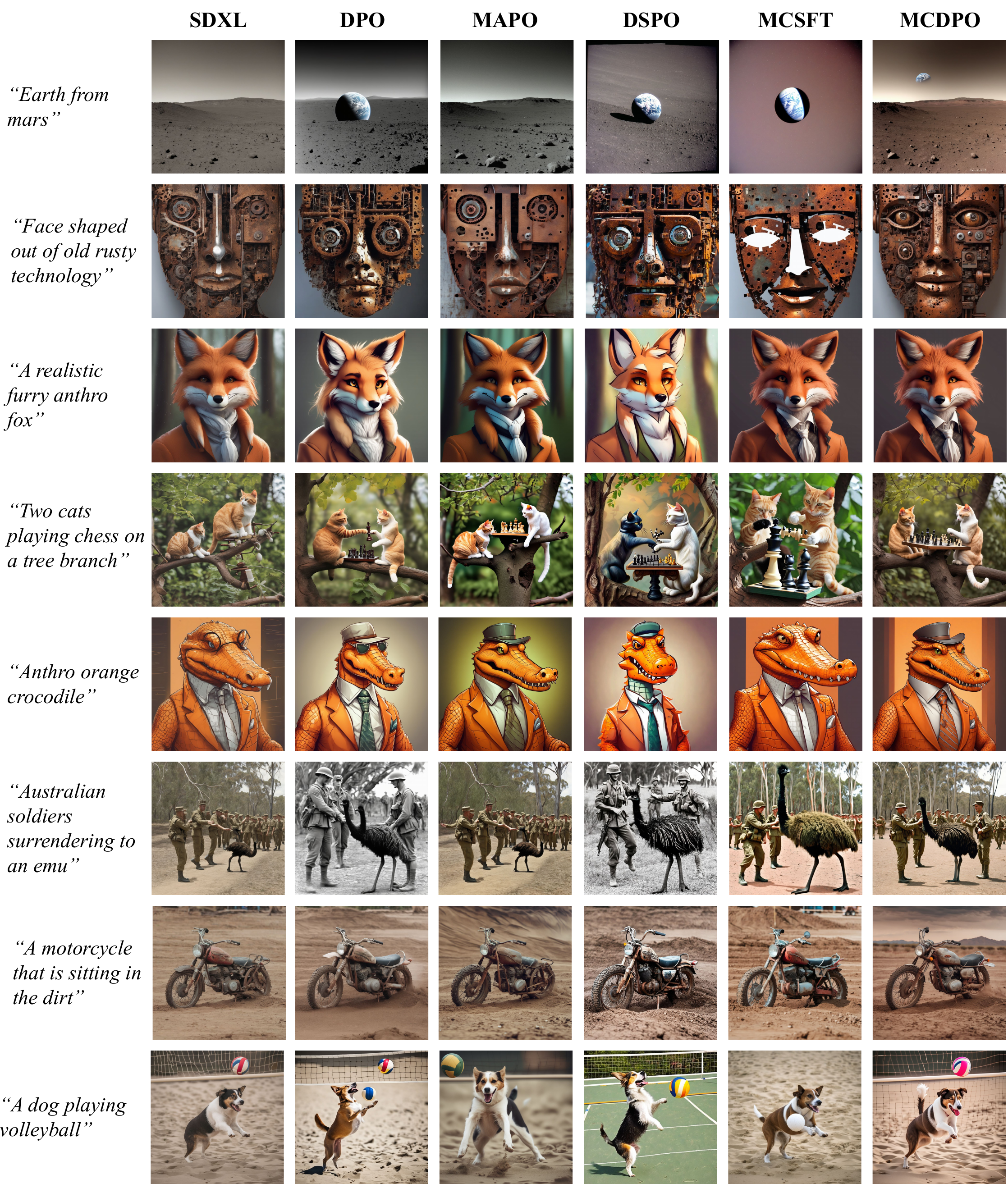}
    \caption{Additional qualitative results of SDXL.}
    \label{fig:sdxl_supp}
\end{figure*}

\end{document}